\documentclass{esannV2}
\usepackage[pdftex]{graphicx}
\usepackage[latin1]{inputenc}
\usepackage{amssymb,amsmath,array}
\usepackage{booktabs}
\usepackage{algorithm}
\usepackage{algpseudocode}
\usepackage{caption}
\usepackage{subcaption}
\usepackage{adjustbox}
\usepackage{hyperref}

\begin{document}
\title{Scalable Forward-Forward Algorithm}

\author{Andrii Krutsylo
%
\vspace{.3cm}\\
%
Polish Academy of Sciences - Institute of Computer Science \\
Jana Kazimierza 5, 01-248, Warsaw - Poland
}

\maketitle

\begin{abstract}

We propose a scalable Forward-Forward (FF) algorithm that eliminates the need for backpropagation by training each layer separately. Unlike backpropagation, FF avoids backward gradients and can be more modular and memory efficient, making it appealing for large networks. We extend FF to modern convolutional architectures, such as MobileNetV3 and ResNet18, by introducing a new way to compute losses for convolutional layers. Experiments show that our method achieves performance comparable to standard backpropagation. Furthermore, when we divide the network into blocks, such as the residual blocks in ResNet, and apply backpropagation only within each block, but not across blocks, our hybrid design tends to outperform backpropagation baselines while maintaining a similar training speed. Finally, we present experiments on small datasets and transfer learning that confirm the adaptability of our method.

\end{abstract}

\section{Introduction}

Artificial neural networks have transformed machine learning, with backpropagation serving as the standard algorithm for training them. Despite its success, backpropagation has well-known drawbacks: it requires differentiable operations in each layer, requires high memory consumption, fixes the network architecture, and suffers from exploding or vanishing gradients. In addition, it provides limited interpretability of the evolution of the representations.

These limitations have motivated research into alternative optimization methods, such as the Forward-Forward (FF) algorithm. Instead of using backward gradients, FF updates each layer with forward passes only, setting local objectives based on positive (correct) and negative (incorrect or corrupted) data. This can reduce memory usage and provide more transparency by allowing each layer to learn independently. However, applying FF to large convolutional architectures is challenging because changes to a few input pixels often do not propagate effectively through deep CNN layers, while significant changes to the entire image can degrade performance.

To overcome these limitations, several methods have been proposed to adapt FF optimization to convolutional layers. These methods typically involve complicated mechanisms for generating positive and negative data, or require architectural modifications that complicate implementation and hinder the practical adoption of FF in large-scale models.

We introduce the Scalable Forward-Forward (SFF) algorithm, the first to work with large convolutional models such as ResNet18 \cite{resnet} and MobileNetV3 \cite{mobilenet}. Our main contributions are the following:

\begin{enumerate}
    \item \textbf{One-Pass Positive/Negative Goodness Computation}: We build on the idea of the Channel-wise Competitive Learning Forward-Forward Algorithm (CwC) \cite{cwc} of extracting both positive and negative goodness in a single forward pass. However, instead of hard assigning channel groups to labels, we add small auxiliary convolutional layers that compute label-specific goodness. This approach removes the class-channel assignment limitations of CwC, maintains low computational overhead, and eliminates the need for architectural changes, replacing complex CwC blocks.
    \item \textbf{Improved Loss Function for Convolutional Layers}: The single forward pass approach requires a different loss function than the original FF. We introduce a log-sum-exp smoothing step to compare positive goodness against all goodness values in a single pass. This improves upon the CwC loss by eliminating the need for probabilistic normalization or channel partitioning. The smoothed comparisons reduce sharp gradient changes, enhancing training stability.
    \item \textbf{Hybrid Block-wise FF + Backpropagation}: To handle deeper architectures, we enable backpropagation within each model block, but not between blocks. This preserves the layerwise advantages of FF while supporting more complex interactions within blocks. We show that this hybrid approach can outperform standard backpropagation, while pure FF remains effective for simpler CNNs.
    \item \textbf{Transfer Learning Between Backpropagation and FF}: We show that networks pretrained with backpropagation can be successfully fine-tuned with FF, suggesting that both optimization methods learn compatible feature hierarchies.
    \item \textbf{Small-Data Problem}: We compare standard backpropagation with our Scalable FF (SFF) on reduced datasets containing only 1,000 random training samples. SFF outperforms backpropagation in this low-data regime, indicating that it may become a preferred solution for small-data problems.
\end{enumerate}

The Scalable Forward-Forward algorithm stands as a direct replacement for backpropagation, delivering several compelling advantages. It frequently achieves superior performance while requiring significantly lower memory allocation, improving efficiency for large-scale applications. Crucially, SFF does not require any modifications to existing model architectures, ensuring seamless compatibility with modern convolutional neural network designs. From an implementation perspective, SFF is straightforward to integrate, acting as a simple wrapper around the model object without introducing additional limitations.

The implementation of the proposed method, including all code and scripts necessary to reproduce the experiments, is available in a publicly accessible GitHub repository at \href{https://github.com/DentonJC/sff}{https://github.com/DentonJC/sff}.

\section{Background and Related Work}

\subsection{The Forward-Forward algorithm}

The Forward-Forward (FF) algorithm \cite{ff} introduces a different approach to neural network training. Instead of relying on the backward pass of backpropagation to compute gradients and update all parameters simultaneously, FF uses two forward passes per training iteration. One forward pass uses \emph{positive} data (i.e., data that should be recognized or correctly classified), while the other uses \emph{negative} data (i.e., data that should not be recognized or that is intentionally corrupted or labeled incorrectly). At the heart of FF is the notion of a \emph{goodness} function within each layer, which measures the desirability of the activations in that layer. In the original algorithm, the goodness function is defined as the sum of the squares of the activations. The training objective in FF is to maximize the goodness of each layer when presented with positive data, and to minimize the goodness when presented with negative data.

In the supervised setting, the construction of positive and negative samples involves embedding label information directly into the input. For example, to create a positive sample for label $y$, we embed this label in the first few pixels (or some designated region) of the input image in such a way that the network can detect the presence of the correct label. This modified image is then passed through each layer in a forward pass, yielding activations $\{ \mathbf{h}_\ell \}$ for layers $\ell = 1,2,\dots,L$. For positive data, the FF algorithm attempts to increase the sum of the squared activations in each layer, effectively pushing the network parameters to produce higher activations whenever the correct label is embedded in the input.

A negative sample in the supervised version can be constructed by taking an image from the dataset and embedding an incorrect label (i.e., any label $y` \neq y$). By forwarding this negatively labeled sample through the same network, the goodness in each layer is computed again. This time, the FF algorithm seeks to reduce the sum of squared activations by pushing the parameters to be less responsive to incorrect input-label pairs. Because each layer's parameters are updated separately to reflect these goodness objectives, the training proceeds in a way that does not require computing backward gradients through the entire network. Instead, each layer can locally update its weights to move the activations in the correct direction without the need for backpropagation through deeper or shallower layers.

While the supervised version injects label information directly into the inputs, an unsupervised variant of the FF algorithm can be used for data without explicit labels. In the unsupervised case, the network is typically exposed to raw data as positive examples and some form of corrupted or artificially generated data as negative examples. The principle remains the same: each layer is trained to produce a higher sum of squared activations for real (positive) data and a lower sum of squared activations for corrupted (negative) data. Over time, the layers learn to distinguish real structure (e.g., edges, shapes, and textures) from artificial or meaningless patterns, thereby capturing features in an unsupervised manner.

In both supervised and unsupervised versions of FF, a major attraction is that the parameters of each layer can be updated in parallel or in a feedforward fashion. However, in the original formulation, the requirement to embed label information in the input data to generate positive and negative samples poses a challenge to modern deep convolutional architectures. Embedding label information in pixel space may disrupt the spatial structure of images or may not influence the activations of convolutional layers enough to compute meaningful goodness.

The Forward-Forward algorithm offers a range of advantages over traditional backpropagation. It is more biologically plausible, eliminating the need for explicit backward passes and error derivative propagation. It also reduces algorithmic complexity, making it easier to implement in heterogeneous or modular systems. Unlike backpropagation, FF does not depend on differentiable models of intermediate operations, allowing it to seamlessly integrate black-box components and non-differentiable transformations. The locally optimized objective of each layer means that the training process is decoupled, allowing layers to be trained independently or in parallel, which can improve scalability and modularity.

These characteristics make FF particularly well suited for low-power or constrained environments, since it does not require storage of intermediate activations or extensive gradient computations.

\subsection{Modifications of the Forward-Forward algorithm}

The spatial-extended labeling method \cite{spatial} is a straightforward approach to overcome the limitations of label embedding in FF optimization.This technique superimposes a second image of the same dimensions on the input image and encodes positive or negative label information using a gray-scale waveform characterized by specific frequency, phase, and orientation parameters unique to each label, which shows good results on small grayscale datasets such as MNIST \cite{mnist} or FashionMNIST \cite{fmnist}.

Another refinement involves the use of the orientation and length of the activation vectors in a multilayer FF optimization process \cite{orientation}. Positive and negative samples are fed into the first layer, where layer normalization computes the orientation and length of the activation vector. The orientation serves as the input to the next layer, while the length contributes to the goodness calculation. Each layer is optimized to assign high goodness to positive samples (above a defined threshold) and low goodness to negative samples (below the threshold). At inference time, the sum of the goodness values across all layers, excluding the first, is computed for each possible label. 

The Self-Contrastive Forward-Forward Algorithm \cite{contrastive} and the Improved Contrastive Forward-Forward Algorithm \cite{gananath2024improved} both adapt principles of self-supervised learning to generate positive and negative examples in an unsupervised manner. This approach eliminates the dependence on explicit label information during training by constructing positive and negative samples by contrasting different transformations of the same input. The FF algorithm then optimizes the positive and negative goodness functions based on these self-constructed contrasts.

\subsection{Channel-wise Competitive Learning Forward-Forward Algorithm}

One of the most notable variants of the Forward-Forward framework is the Channel-wise Competitive Learning Forward-Forward Algorithm (CwC) \cite{cwc}. Unlike standard methods that require explicit negative samples or label embedding, CwC allows both positive and negative goodness values to be obtained in a single forward pass. It does this by partitioning the feature maps within each layer into channel groups, each corresponding to a particular class. The model then computes positive goodness for the channels associated with the ground truth class and negative goodness for the remaining channels. This design removes the need to create class-specific negative data and offers a streamlined training process.  

\subsubsection{Channel Partitioning and Goodness Computation}
Suppose a convolutional layer produces a tensor
\(
  \mathbf{Y} \in \mathbb{R}^{B \times C \times H \times W}
\)
for a batch of \(B\) samples. The CwC approach divides the \(C\) output channels evenly among \(J\) classes, allocating \(\tfrac{C}{J}\) channels to each class \(j\). Denote by \(\hat{\mathbf{Y}}_{j}\) the sub-block of \(\mathbf{Y}\) for class~\(j\). The corresponding \emph{goodness} factor of class \(j\) for sample \(n\) is
\[
   G_{n,j}
   \;=\;
   \frac{1}{\tfrac{C}{J}\,H\,W}
   \sum_{c=1}^{\tfrac{C}{J}} \sum_{h=1}^{H}\sum_{w=1}^{W} 
   \bigl(\hat{Y}_{n,j,c,h,w}\bigr)^2,
\]
forming a goodness matrix 
\(
  \mathbf{G}\in \mathbb{R}^{B\times J}.
\)
A binary mask 
\(
  \mathbf{Z}\in\{0,1\}^{B\times J}
\)
selects the \emph{positive} (ground-truth) class while marking the others as \emph{negative}. Hence, for each sample~\(n\),
\[
  \mathbf{g}^+ \;=\; \mathbf{G}\,\mathbf{Z}^{\top},
  \quad
  \mathbf{g}^- \;=\; \mathbf{G}\,(\mathbf{1}-\mathbf{Z}^{\top}),
\]
yielding the sample's positive goodness versus the average goodness over non-target classes.

\subsubsection{Softmax-Based CwC Loss}
CwC adopts a softmax-driven competitive loss that encourages the ground-truth class to dominate:
\[
  L_{\mathrm{CwC}}
  \;=\;
  -\frac{1}{B}\sum_{n=1}^B
  \log\!\Bigl(\frac{\exp(g_{n}^+)}{\sum_{j=1}^{J} \exp(G_{n,j})}\Bigr).
\]
This formulation focuses on maximizing the exponentiated goodness of the correct class relative to all other classes.

\subsubsection{Channel-wise Feature Separator and Extractor}
To improve class-specific feature learning, CwC introduces the Channel-wise Feature Separator and Extractor (CFSE) block. Each CFSE block consists of:
\begin{enumerate}
    \item A \textbf{standard convolution sub-block} that processes the entire feature space to capture shared and inter-class features.
    \item A \textbf{grouped convolution sub-block} that partitions the feature space by channel, allowing subsets of channels to focus on class-specific signals.
\end{enumerate}
Each sub-block uses ReLU activation, batch normalization, and optional max pooling to stabilize training and improve feature discrimination. This design aims to balance intra-class feature extraction with reduced parameter complexity.

\subsubsection{Interleaved Layer Training (ILT)}
Finally, CwC employs an \emph{Interleaved Layer Training (ILT)} strategy. Rather than training each layer in isolation or freezing earlier layers prematurely, as proposed in the original FF, ILT activates layer-specific training epochs in a staggered fashion. The method begins by identifying the appropriate starting and stopping epochs. It trains the first layer until its performance plateaus, marking this as the stopping epoch. For each subsequent layer, the training is restarted, and their respective plateau points and stopping epochs are recorded.

In the second phase, ILT conducts the actual training of the model using the recorded epochs. It performs interleaved training where all layers are trained together. However, instead of training continuously, each layer stops at its previously determined plateau epoch. This approach helps prevent the model from getting stuck in local minima and ensures that each layer is fine-tuned based on the consistent outputs from the previous layer. This interleaving helps layers avoid local minima and adapt even after subsequent layers have begun learning more refined representations.

\subsubsection{Practical Constraints}
Despite its advantages, CwC often requires significant modification of existing models to accommodate CFSE blocks. Partitioning channels by class can become unreliable for large problems (e.g., with many classes). For example, in the CIFAR-100 setting, it was necessary to introduce the smaller number of superclasses to train the early layers, and then add an additional classification block for all 100 classes to achieve feasible training. Finally, in its current state, the ILT requires several iterations to train the full model before the optimal schedule for freezing layers can be determined.

\section{Experimental Setup}

\begin{table}[ht!]
\centering
\begin{tabular}{ll}
\toprule
\textbf{Model} & \textbf{Layers/Blocks} \\
\midrule
\textbf{CNN} & 
\begin{tabular}{@{}l@{}} 
- convolutional layer with ReLU \\ 
- pooling layer \\ 
- convolutional layer with ReLU \\ 
- pooling layer \\ 
- convolutional layer with ReLU \\  
- dropout layer \\ 
- linear classifier
\end{tabular}
 \\
\midrule
\textbf{CNNB} &
\begin{tabular}{@{}l@{}} 
- Block 1: \\ 
\quad \(\;\) convolutional layer + BN + ReLU + pooling layer \\
- Block 2: \\ 
\quad \(\;\) convolutional layer + BN + ReLU + pooling layer \\
- Block 3: \\ 
\quad \(\;\) convolutional layer + BN + ReLU \\
- dropout layer \\
- linear classifier
\end{tabular}
 \\
\midrule
\textbf{ResNet18} & 
\begin{tabular}{@{}l@{}} 
- initial \(3\times3\) convolutional layer (or \(7\times7\) for Imagenette) \\ 
- 4 groups of residual blocks (each group treated as block) \\ 
- global average pooling \\ 
- linear classifier
\end{tabular}
 \\
\midrule
\textbf{MobileNetV3} &
\begin{tabular}{@{}l@{}} 
- initial \(3\times3\) convolutional layer (or \(7\times7\) for Imagenette) \\ 
- 11 inverted residual blocks (the small version) \\ 
- final convolutional layer (not used for SFF) \\
- classifier head
\end{tabular}
 \\
\bottomrule
\end{tabular}
\caption{Comparison of model architectures. For CIFAR-10 and CIFAR-100, ResNet18 and MobileNetV3 have their first convolutional layer kernel size adjusted to \(3\times3\). The batch normalization (BN) layer should always be in block, otherwise it will decrease the model performance. In the context of SFF, \emph{block} means that only the outputs of the last layer of the set would be used for the goodness computations, and layer normalization would be applied after that layer.}
\label{tab:model_comparison}
\end{table}

We evaluated our methods on the CIFAR-10, CIFAR-100 \cite{cifar}, and Imagenette \cite{imagenette} datasets (Imagenette resized to \(224 \times 224\) pixels). Each dataset was split into training and test sets according to the standard protocol, and we further randomly divided the training set into 80\% training and 20\% validation for each random seed.

Table \ref{tab:model_comparison} summarizes the architectures used in our experiments: CNN, CNNB, ResNet18, and MobileNetV3. For ResNet18 and MobileNetV3, we reduced the kernel size of the first convolutional layer to \(3 \times 3\) on CIFAR-10 and CIFAR-100 to better accommodate smaller images; for Imagenette, we left it unchanged. The CNN and CNNB models share the same layers, except that CNNB combines the convolution, batch normalization, and pooling layers into a single block, while CNN has no batch normalization layers. Under Forward-Forward optimization (both SFF and CwC), each block is processed with forward passes, allowing backpropagation within that block.

To stabilize training under FF, we applied layer normalization after each trainable layer (or block), following the original FF guidelines. As a baseline, we trained equivalent models using standard backpropagation and cross-entropy loss, omitting the additional normalization layers. We used the AdamW optimizer and performed hyperparameter searches for two learning rates (one for all feature extraction layers, one for the classifier) and weight decay. The learning rates tested were in the set $\{7\times10^{-5},\, 1\times10^{-4},\, 3\times10^{-4},\, 5\times10^{-4},\, 7\times10^{-4},\, 1\times10^{-3},\, 3\times10^{-3},\, 5\times10^{-3},\, 7\times10^{-3},\, 1\times10^{-2},\, 3\times10^{-2},\, 5\times10^{-2},\, 7\times10^{-2},\, 1\times10^{-1},\, 3\times10^{-1},\, 5\times10^{-1}\}$, while the weight decay was selected from $\{{0.0}, 10^{-8}, 10^{-7}, \\ 10^{-6}, 10^{-5}\}.$

Whenever the validation loss did not improve for 10 epochs, the learning rate was reduced, for each layer independently in the case of FF optimization. Batch sizes were set to 128 for CIFAR-10 and CIFAR-100, and 64 for Imagenette. We trained each network for up to 100 epochs, using early stopping if the validation accuracy did not improve for 30 consecutive epochs. For the backpropagation experiments, the number of 100 epochs was never reached.

We repeated each experiment 3 times (random seeds 0 to 2). All experiments were performed on an NVIDIA GeForce GTX 1050 Ti with 4 GB of memory.

\section{Method}

We introduce the Scalable Forward-Forward approach to overcome the limitations of previous FF methods when applied to modern convolutional models and datasets with a large number of classes, making it a ready-to-use replacement for backpropagation with minimal hyperparameter tuning. The only tunable parameter in this approach is the kernel size of the auxiliary convolutional layer.

\subsection{Layerwise Goodness Computation}

Let each layer map an input tensor $\mathbf{X}\in\mathbb{R}^{B\times C\times H\times W}$ into an output $\mathbf{Y}\in\mathbb{R}^{B\times D\times H'\times W'}$. For a given layer, we further transform $\mathbf{Y}$ with a small convolution kernel of size $1\times1$ (or slightly larger) to produce a goodness tensor $\mathbf{Z}\in\mathbb{R}^{B\times J\times H'\times W'}$, where $J$ is the number of classes. Averaging or reshaping $\mathbf{Z}$ yields a class-specific vector of goodness values,
\[
  \mathrm{gf}_{n,j} \;=\; f(\mathbf{Z})_{n,j}\quad\text{for}\quad j = 1,\dots,J,
\]
where $f(\cdot)$ is a simple pooling operator (e.g., average over spatial dimensions).

\subsection{Objective Function}

To train layerwise, we adopt a margin-based objective reminiscent of log-sum-exp margins:
\[
   L_{\mathrm{SFF}} 
   \;=\; -\,\mathbb{E}\Bigl[
         g_{\mathrm{pos}}
         \;-\;
         \log\!\!\sum_{j=1}^{J}\exp\bigl(\mathrm{gf}_{n,j}\bigr)
        \Bigr],
\]
where $g_{\mathrm{pos}}$ is the goodness corresponding to the ground-truth class. This formulation encourages the correct class to surpass the aggregate of all class scores and obviates the need for additional negative samples. Unlike strictly partitioned approaches, SFF relies on learnable filters to discover the most effective alignment of features to classes.

\subsection{Training}

Here are the training steps of the algorithm for each minibatch and for each layer that has trainable parameters (otherwise, only pass the outputs of the layer to the next layer): 
\begin{enumerate} 
    \item Pass the output of the layer to the next layer (the next layer could be trained in parallel). 
    \item Compute the goodness of the layer, defined as the mean of the squares of the layer's output and the auxiliary convolutional layer on top of it. This should produce a tensor with shape [minibatch size, number of classes]. 
    \item Compute positive and negative goodness factors. The positive factor is the value in the goodness tensor that corresponds to the positive class, while the negative factor is the mean of the remaining classes. (Note: the negative factor is not used in the loss function of the presented results.) 
    \item Compute the local loss $L_{\mathrm{SFF}}^{(\ell)}$ using the positive goodness and the full goodness tensor. 
    \item Use the local loss to perform a parameter update of the current layer and the corresponding auxiliary convolutional layer, whose output is used only for the computation of the local loss and will not be passed to the next layer. 
\end{enumerate}

To simplify and generalize our implementation, making it applicable to any PyTorch model using a simple wrapper class, we allow backpropagation within the blocks of the mode (if the layers are divided into blocks) but restrict it between blocks. This approach allows us, for example, to take advantage of the residual blocks in ResNet internally while externally adhering to FF optimization and retaining all its benefits.

\subsection{Evaluation}

After each layer is independently updated, we can use its local goodness tensor for classification by selecting the index of the highest value as the predicted class. For the reported results, we average the goodness vectors from all layers to make predictions, since the goodness or activations of the last layer only show lower accuracy. This is a known issue in layerwise training. The typical way to address it is to pass the aggregated activations of all layers to the final classification head, which is trained via a standard supervised loss, such as cross-entropy. However, this approach would require substantial modification of the model and increase memory consumption, whereas the main strength of SFF is its simplicity and memory efficiency as a ready-to-use replacement for backpropagation.

\section{Results}

\begin{table}[htbp]
\centering
\begin{tabular}{lccccc}
\toprule
\multicolumn{6}{c}{\textbf{CIFAR-10 - CNN}} \\
\toprule
& \multicolumn{1}{c}{\textbf{Layers Acc}} & \multicolumn{1}{c}{\textbf{Head Acc}} & \textbf{VRAM} (MB)& \multicolumn{1}{c}{\textbf{Time}} & \multicolumn{1}{c}{\textbf{\# Params}} \\
\midrule
BP& -                & 79.10 $\pm$ 0.49& 544.91& 1   & 175178\\
CwC-mod     & 69.06 $\pm$ 1.11& 69.60 $\pm$ 0.24& 579.23& 3& 175626\\
SFF              & 70.34 $\pm$ 0.05& -                & 537.34& 2.6& 149726\\

\toprule
\multicolumn{6}{c}{\textbf{CIFAR-10 - CNNB}} \\
\toprule
& \multicolumn{1}{c}{\textbf{Layers Acc}} & \multicolumn{1}{c}{\textbf{Head Acc}} & \textbf{VRAM} & \multicolumn{1}{c}{\textbf{Time}} & \multicolumn{1}{c}{\textbf{\# Params}} \\
\midrule
BP& - & 78.49 $\pm$ 0.22& 544.93& 1 & 175626\\
CwC-mod          & 72.88 $\pm$ 0.25& 76.17 $\pm$ 0.63& 579.26& 2.7& 176074\\
SFF           & 81.38 $\pm$ 0.35& - & 528.3& 2.2& 150174\\

\toprule
\multicolumn{6}{c}{\textbf{CIFAR-100 - CNN}} \\
\toprule
& \multicolumn{1}{c}{\textbf{Layers Acc}} & \multicolumn{1}{c}{\textbf{Head Acc}} & \textbf{VRAM} & \multicolumn{1}{c}{\textbf{Time}} & \multicolumn{1}{c}{\textbf{\# Params}} \\
\midrule
BP& - & 45.64 $\pm$ 0.52& 561.85& 1 & 912548\\
CwC-mod     & 1.00 $\pm$ 0.00& 1.00 $\pm$ 0.00& 598.09& 2& 912996\\
SFF           & 45.79 $\pm$ 0.53& - & 883.78& 7.4& 295596\\

\toprule
\multicolumn{6}{c}{\textbf{CIFAR-100 - CNNB}} \\
\toprule
& \multicolumn{1}{c}{\textbf{Layers Acc}} & \multicolumn{1}{c}{\textbf{Head Acc}} & \textbf{VRAM} & \multicolumn{1}{c}{\textbf{Time}} & \multicolumn{1}{c}{\textbf{\# Params}} \\
\midrule
BP& - & 50.49 $\pm$ 0.36& 561.12& 1 & 912996\\
CwC-mod     & 1.00 $\pm$ 0.00& 1.00 $\pm$ 0.00& 598.11& 1.5& 913444\\
SFF           & 55.34 $\pm$ 0.07& - & 702.08& 4.2& 296044\\

\toprule
\multicolumn{6}{c}{\textbf{Imagenette - MobileNetV3}} \\
\toprule
& \multicolumn{1}{c}{\textbf{Layers Acc}} & \multicolumn{1}{c}{\textbf{Head Acc}} & \textbf{VRAM} & \multicolumn{1}{c}{\textbf{Time}} & \multicolumn{1}{c}{\textbf{\# Params}} \\
\midrule
BP& - & 80.44 $\pm$ 0.39& 1048.79& 1 & 932778\\
CwC-mod          & 74.56 $\pm$ 0.48& 72.41 $\pm$ 0.38& 373.71& 2.1& 981306\\
SFF           & 77.06 $\pm$ 0.44& - & 516.98& 1.4 & 1017848\\

\toprule
\multicolumn{6}{c}{\textbf{Imagenette - ResNet18}} \\
\toprule
& \multicolumn{1}{c}{\textbf{Layers Acc}} & \multicolumn{1}{c}{\textbf{Head Acc}} & \textbf{VRAM} & \multicolumn{1}{c}{\textbf{Time}} & \multicolumn{1}{c}{\textbf{\# Params}} \\
\midrule
BP& - & 81.84 $\pm$ 0.45& 1675.47& 1 & 11702322\\
CwC-mod        & 83.97 $\pm$ 0.82& 82.68 $\pm$ 0.56& 1058.39& 1.6& 11183690\\
SFF           & 83.24 $\pm$ 0.44& - & 1393.38& 1.2& 11143610\\

\bottomrule 
\end{tabular}
\caption{Classification performance of backpropagation (BP), our modified Channel-wise Competitive Learning (CwC-mod), and the proposed Scalable Forward-Forward (SFF) across multiple architectures and datasets. For each dataset, we report the layer-ensemble accuracy (Layers Acc), the final head accuracy (Head Acc), peak VRAM usage, wall-clock time relative to BP (Time), and the total number of parameters (Params). The layer accuracy aggregates predictions from feature-extraction layers or blocks, whereas the head accuracy is produced by the classifier head. Peak VRAM usage was measured using PyTorch's \emph{max\_memory\_allocated}, and the time is an approximate ratio to BP's total runtime. The parameter counts vary because SFF includes auxiliary layers but does not rely on the standard classifier head.}
\label{tab:results}
\end{table}

To evaluate the effectiveness of our proposed method, we compared the classification performance of backpropagation (BP), the modified version of Channel-wise Competitive Learning (CwC-modified), and our Scalable Forward-Forward (SFF) optimization, as summarized in Table \ref{tab:results}. Since the original CwC lags behind BP by an average of 6.83\% for CIFAR-10 and CIFAR-100 on the model constructed specifically for this method according to \cite{cwc}, we introduced several modifications to make it competitive. Instead of the ILT schedule tailored to each dataset, we use the reduce on plateau scheduler for each individual layer or block. There are no CFSE blocks in the models we use for experiments, so we extract the goodness directly from the layers or blocks, but the goodness and loss functions remain as in the original CwC. Finally, we allow backpropagation within the blocks of the CNNB, MobilenetV3, and ResNet18 models for both CwC and SFF, while the CNN model only has layers that are not grouped into blocks. We do not consider other layerwise methods, since they are either not comparable to BP in performance and speed, or have not yet been applied to modern convolutional models.

From the SFF perspective, the main difference between the tested models is the size and number of blocks, with a clear benefit for efficiency from the larger number of smaller blocks (MobileNetV3). Splitting the model into more blocks, while requiring the individual auxiliary layers, increases the number of parameters, potentially simplifying the training of each block. The smaller the block, the smaller the maximum VRAM consumption, because while we need to store the entire model weights in VRAM for faster computations (which is optional for SFF), we also need to store the activations and gradients for the current block. MobileNetV3 blocks are smaller than in ResNet18, which is reflected in the difference in VRAM consumption between BP and SFF, but does not affect the number of parameters. 
In all experimental settings, SFF is slower than BP, and the difference is most pronounced for the large number of classes. Partly this is due to the larger convolutional auxiliary layers required, but the main problem may be in the loss function required from each individual block to learn the difference between all classes. Making the earlier layers to learn the general features could improve the training speed and performance.
The performance comparison shows that for all experiments, except for the CNN model, SFF is similar or better than BP. The modified CwC method also shows similar performance in most cases, but is generally slower and cannot handle the large number of classes. The difference in SFF results between CNN and CNNB shows that the pure FF approach is possible, but the backpropagation blocks have a much higher usability. 

\begin{table}[htbp]
\centering
\begin{tabular}{lllc}
\toprule
\multicolumn{1}{c}{\textbf{Dataset}} & \multicolumn{1}{c}{\textbf{Model}} & \multicolumn{1}{c}{\textbf{Method}} & \multicolumn{1}{c}{\textbf{Accuracy}} \\
\midrule
CIFAR-10& CNNB& BP& 49.90 $\pm$ 0.61\\
CIFAR-10& CNNB& SFF& 50.19 $\pm$ 1.04\\
Imagenette& MobileNetV3& BP& 49.93 $\pm$ 2.04\\
 Imagenette& MobileNetV3& SFF& 58.95 $\pm$ 1.62\\ 
\bottomrule
\end{tabular}
\caption{Classification accuracy of BP and SFF under limited training data (1000 samples). The reported mean and standard deviation are averaged over three runs.}
\label{tab:results_small}
\end{table}

We simulated the small data setting by randomly selecting only 1000 samples and compared the performance of BP and SFF in Table \ref{tab:results_small}. While for the small convolutional model SFF shows similar performance to BP, for the large model the 9\% difference indicates the clear superiority of the SFF method and the need for further investigation. 

\begin{table}[htbp]
\centering
\begin{tabular}{llcc}
\toprule
 \multicolumn{1}{c}{\textbf{Model}} & \multicolumn{1}{c}{\textbf{Method}} & \multicolumn{1}{c}{\textbf{Accuracy}} & \textbf{Improvement}\\
\midrule
 MobileNetV3& BP& 85.14 $\pm$ 1.50 &+4.70\\
 MobileNetV3& SFF& 79.80 $\pm$ 0.71&+2.74\\
 ResNet18& BP& 91.77 $\pm$ 1.45 &+9.93\\
  ResNet18& SFF& 90.73 $\pm$ 0.97 &+7.49\\ 
\bottomrule
\end{tabular}
\caption{Performance improvements brought by pretraining on ImageNet. The \emph{Improvement} column indicates the gain in final accuracy relative to the non-pretrained baseline of each respective method in Table~\ref{tab:results}.}
\label{tab:results_pretrained}
\end{table}

As expected, pretraining with backpropagation and the ImageNet dataset consistently improves the performance of the MobileNetV3 and ResNet18 models trained with backpropagation (Table \ref{tab:results_pretrained}). The interesting result of these experiments is that SFF also benefits significantly from the same pretraining. This suggests that both optimization approaches learn a similar feature hierarchy, which greatly expands the applicability of the SFF method.

\section{Conclusions}

In this work, we introduced the Scalable Forward-Forward algorithm as a drop-in replacement for backpropagation, aimed at modern convolutional neural networks with a large number of classes. Our experiments on CIFAR-10, CIFAR-100, and Imagenette demonstrated that SFF achieves performance comparable to, and in some cases surpassing, conventional backpropagation. Notably, SFF remains effective even for deeper architectures such as ResNet18 and MobileNetV3, without requiring the specialized channel partitioning or complex negative sample generation characteristic of earlier Forward-Forward methods like Channel-wise Competitive Learning.

A key advantage of our approach lies in its \emph{block-wise} design, which combines the layerwise benefits of FF optimization, such as the better explainability or the ability to integrate non-differentiable blocks, with partial backpropagation within local blocks. This strategy preserves the modularity and reduced memory requirements of FF, while allowing deeper layers to learn more sophisticated feature interactions. We also found that pretraining on ImageNet not only improves the performance of standard backpropagation, but also confers significant gains to SFF, indicating that both methods learn compatible feature hierarchies and that FF-trained weights are amenable to transfer learning.

Furthermore, the results in the small data regime show that SFF can excel where backpropagation often suffers from overfitting. Specifically, on Imagenette with only 1000 samples, SFF significantly outperformed backpropagation, suggesting that layerwise training may foster more robust, generalizable representations in small data settings. Future work will explore this property in more detail and investigate whether SFF's resilience to data scarcity can be exploited in diverse domains such as medical imaging, remote sensing, and other tasks with naturally constrained datasets.

Despite these promising findings, our experimental analyses also reveal new directions for improvement. The reliance on auxiliary convolutional layers adds additional parameters and, in some architectures, increased training time. Methods to reduce this overhead, such as parameter sharing, more efficient pooling strategies, or pruning redundant class-specific filters, are promising directions for further optimization.


\begin{footnotesize}

\bibliographystyle{unsrt}
\bibliography{esann.bib}

\end{footnotesize}


\end{document}